\documentclass[letterpaper, 10 pt, conference]{ieeeconf}
\IEEEoverridecommandlockouts
\usepackage[space, compress, sort]{cite}

\usepackage{amsmath, amssymb, amsfonts, mathtools, amsthm}
\usepackage[ruled,vlined]{algorithm2e}
\usepackage{tabularx}
\usepackage{graphicx}
\usepackage{enumerate}
\usepackage{float}
\usepackage{url}
\usepackage{verbatim}
\usepackage{booktabs} 
\usepackage{multirow}
\usepackage[linkcolor=black,citecolor=black,urlcolor=black,colorlinks=true]{hyperref}
\usepackage{bm}
\usepackage{subfigure}
\usepackage{algpseudocode}
\usepackage{xcolor}
\usepackage{tcolorbox}
\newcommand{\tabincell}[2]{\begin{tabular}{@{}#1@{}}#2\end{tabular}}

\definecolor{lightgray}{rgb}{0.96, 0.96, 0.98}
\definecolor{lightyellow}{rgb}{0.98, 0.97, 0.97}
\newcommand{\llmbox}[1]{%
\begin{tcolorbox}[
  colback=lightgray,
  colframe=white,
  left=1.7mm,
  right=1.7mm,
  top=0mm,
  bottom=0mm,
  boxrule=0.0pt,
  boxsep=3pt,
  before skip=3pt,
  after skip=3pt
]
\fontsize{9pt}{9.5pt}\selectfont\ttfamily #1
\end{tcolorbox}
}
\newcommand{\llmoutput}[1]{%
\begin{tcolorbox}[
  colback=lightyellow,
  colframe=white,
  left=1.7mm,
  right=1.7mm,
  top=0.2mm,
  bottom=0mm,
  boxrule=0.0pt,
  boxsep=3pt,
  before skip=3pt,
  after skip=3pt
]
\fontsize{9pt}{9.5pt}\selectfont\ttfamily #1
\end{tcolorbox}
}

\newtcolorbox{mycolorbox}[1][]{commonstyle,#1}

\title{ \LARGE \bf Hierarchical LLMs In-the-Loop Optimization for Real-Time Multi-Robot Target Tracking under Unknown Hazards}

\author{Yuwei Wu$^{1}$, Yuezhan Tao$^{1}$, Peihan Li$^{2}$, Guangyao Shi$^{3}$, \\Gaurav S. Sukhatme$^{3}$, Vijay Kumar$^{1}$,  Lifeng Zhou$^{2}$\textsuperscript{\textdagger} 
\thanks{$^{1}$Yuwei Wu, Yuezhan Tao, and Vijay Kumar are with the GRASP Lab, University of Pennsylvania, Philadelphia, PA 19104, USA. Email: \texttt{\small \{yuweiwu, yztao, kumar\}@seas.upenn.edu}.
$^{2}$Peihan Li and Lifeng Zhou are with the Department of Electrical and Computer Engineering, Drexel University, Philadelphia, PA 19104, USA. Email: \texttt{\small \{pl525,lz457\}@drexel.edu}. 
$^{3}$Guangyao Shi and Gaurav S. Sukhatme are with the Department of Computer Science, University of Southern California, Los Angeles, CA 90089, USA. Email:
\texttt{\small \{shig, gaurav\}@usc.edu}. \textsuperscript{\textdagger} Corresponding author. } 
\thanks{This research was sponsored by the ARL DCIST CRA W911NF-17-2-0181 and the TILOS under NSF grants CCR-2112665.}
}

\begin{document}
\maketitle
%%%%%%%%%%%%%%%%%%%%%%%%%%%%%%%%%%%%%%%%%%%%%%%%%%%%%%%%%%%%%%%%%%%%%%%%%%%%%%%%
\begin{abstract}
Real-time multi-robot coordination in hazardous and adversarial environments requires fast, reliable adaptation to dynamic threats.
While Large Language Models (LLMs) offer strong high-level reasoning capabilities, the lack of safety guarantees limits their direct use in critical decision-making.
In this paper, we propose a hierarchical optimization framework that integrates LLMs into the decision loop for multi-robot target tracking in dynamic and hazardous environments.
Rather than generating control actions directly, LLMs are used to generate task configuration and adjust parameters in a bi-level task allocation and planning problem. 
We formulate multi-robot coordination for tracking tasks as a bi-level optimization problem, with LLMs to reason about potential hazards in the environment and the status of the robot team and modify both the inner and outer levels of the optimization.
This hierarchical approach enables real-time adjustments to the robots' behavior. 
Additionally, a human supervisor can offer broad guidance and assessments to address unexpected dangers, model mismatches, and performance issues arising from local minima.
We validate our proposed framework in both simulation and real-world experiments with comprehensive evaluations, demonstrating its effectiveness and showcasing its capability for safe LLM integration with multi-robot systems. 

\end{abstract}

\IEEEpeerreviewmaketitle
%%%%%%%%%%%%%%%%%%%%%%%%%%%%%%%%%%%%%%%%%%%%%%%%%%%%%%%%%%%%%%%%%%%%%%%%%%%%%%%%
\section{Introduction}

%%%%%%%%%%%%%%%%
Multi-robot coordination and planning have been extensively employed in different applications like exploration, search and rescue, and target tracking~\cite{1435481, ma2018multi, shi2021communication, BOLOURIAN2020103250, 9349130}.
Coordinating a team of robots in unknown and adversarial environments becomes significantly challenging due to the increased complexity and unpredictability of conditions.
Moreover, sensor and communication failures, potentially caused by adversaries, can lead to critical safety issues, including the inability of robots to localize, observe their surroundings, and maintain team connectivity.
While existing approaches have focused on designing algorithms to handle modeled adversarial tasks and environments~\cite{liu2024multi, li2024resilienct, zhou2022distributed, liu2022decentralized}, the mismatch between these models and the unexpected dangers encountered in real-world scenarios makes adaptation difficult.

\begin{figure}[!t]
    \centering
    \includegraphics[width=1\columnwidth]{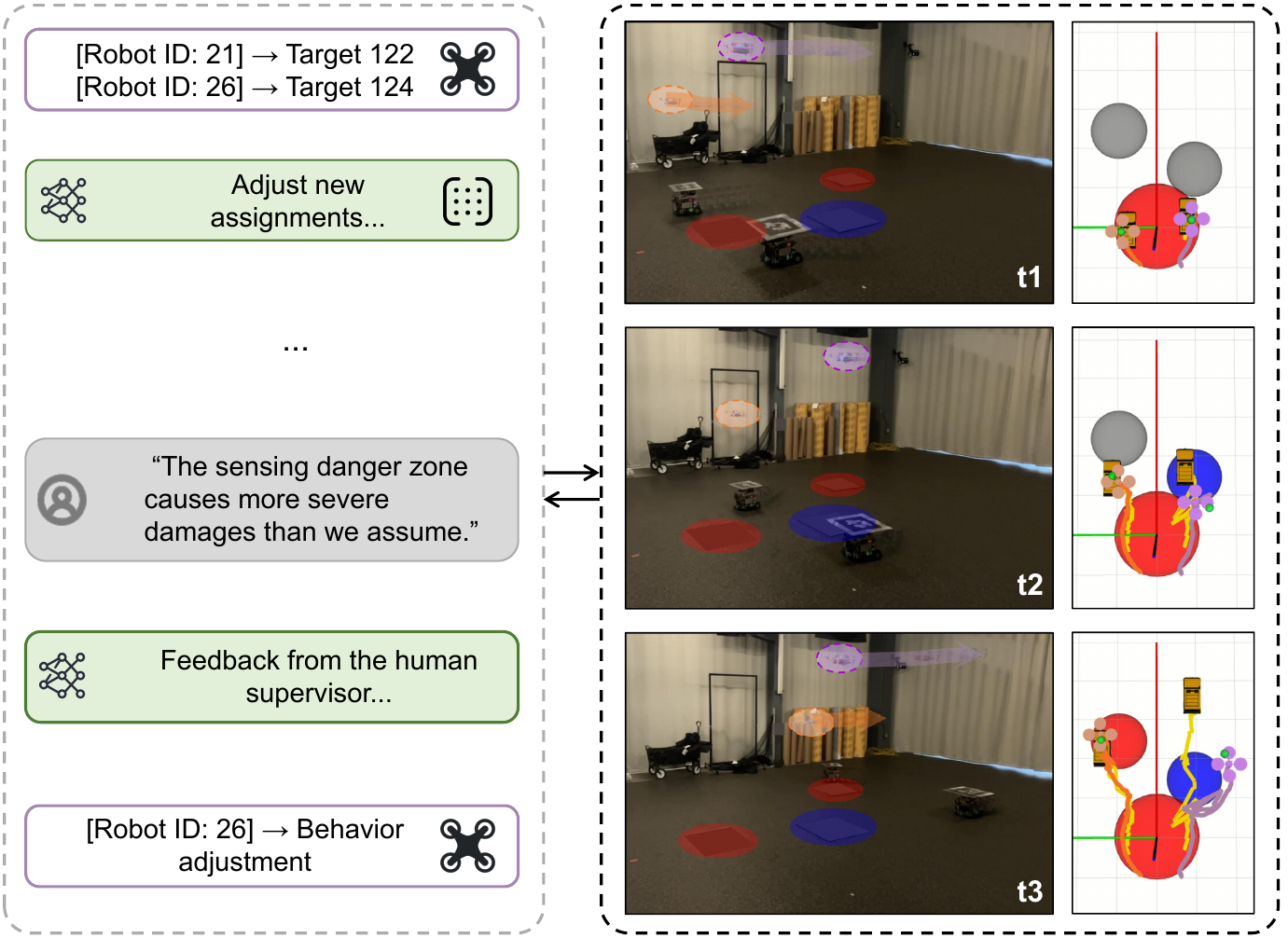}
    \caption{Real-world experiments. Two drones track ground robots in an environment with two sensing zones (red) and one communication zone (blue), guided by human supervision to mitigate risks.}
    \label{fig: fig1}
    \vspace{-0.5cm}
\end{figure}

%%%%%%%%%%%%%%%%%%%%%%%%%%%%%%%%%%%
The recent advancements in Large Language Models (LLMs) have demonstrated great potential for improving task and motion planning for robots~\cite{yu2023co, 10759765, kannan2023smart, sinha2024real, 10801327, chen2024scalable}.
Existing approaches primarily focus on using LLMs for offline applications to provide high-level guidance and rewards for sequential tasks without strict temporal constraints or real-time requirements.
For high-level tasks, LLM-based systems typically depend on human experts to offer linguistic or numerical guidance, refining strategies that enable the robot team to navigate complex and uncertain environments effectively~\cite{shek2023lancar}.
However, mismatches between models and real-world conditions, along with unexpected risks, in multi-robot target tracking can overwhelm human experts due to excessive cognitive demands~\cite{rosenfeld2017intelligent, peters2015human}.
% and humans may not react fast enough.
Moreover, in many practical applications, it is more common for non-expert human supervisors to watch over the robot team. 
These supervisors can offer vague natural language suggestions but typically lack the expertise to adjust mathematical models~\cite{brawer2023towards, shi2024inverse, wakayama2023probabilistic}.
This highlights the need for a \textbf{real-time} coordination assistant to continuously monitor the robot team, incorporate human suggestions, and make strategic adaptations for decision-making.

\begin{figure*}[!t]
    \centering
    \includegraphics[width=2\columnwidth]{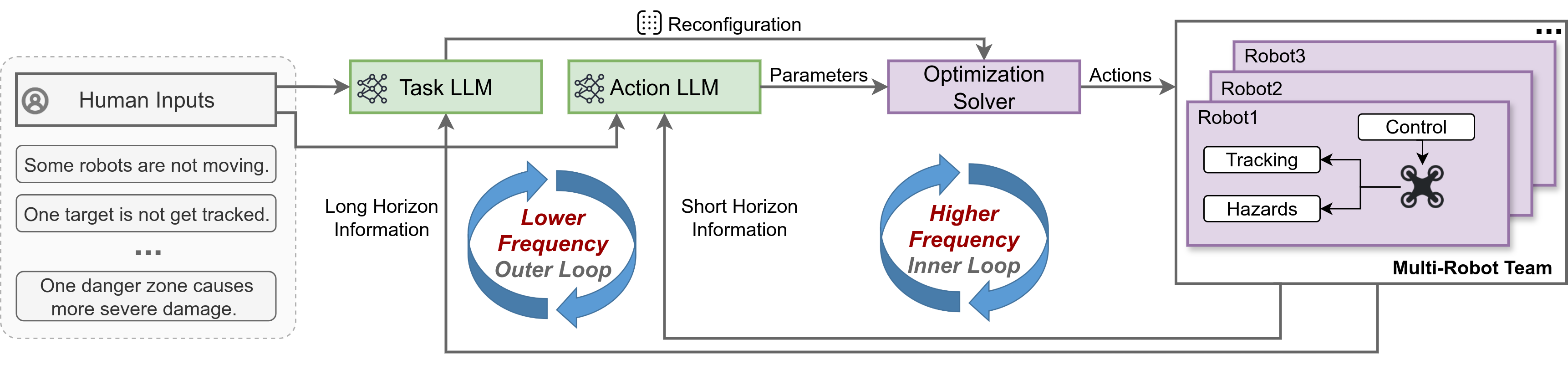}
    \caption{The hierarchical LLM framework for multi-robot tracking processes both robot and human inputs through two LLMs: a high-frequency action LLM for reactive performance adaptation, and a low-frequency task LLM to provide strategic guidance and reconfiguration to the system. The optimization solver takes the LLMs' outputs into executable actions, guiding the multi-robot team in real-time. 
    }
    \label{fig: system}
    \vspace{-0.4cm}
\end{figure*}

To address these challenges, we propose a hierarchical framework that combines the reasoning capabilities of LLMs with optimizers, enabling real-time, robust multi-robot coordination in hazardous environments.
We formulate the task allocation and planning for target tracking under communication and sensing attacks as a bi-level optimization, and employ a hierarchical LLM architecture to handle reasoning tasks at different levels.
The outer loop LLM identifies task allocation variables to decide which target to track with reasoning significance, while the inner loop LLM adjusts parameters to balance performance, safety, and energy efficiency during execution.
% The inner LLM adjusts parameters to prioritize various objectives, including performance, safety, and energy efficiency, while the outer LLM handles online task allocation for team reconfiguration. 
In addition to numerical feedback, we also enable human interactions in the system to address abnormal behaviors, remind of unexpected danger avoidance, and provide more high-level supervision.
We validate the proposed framework through diverse simulated scenarios and real-world hardware experiments (Fig.~\ref{fig: fig1}), demonstrating its real-time robustness and adaptability.
The contributions of the paper are as follows:
\begin{itemize}
    \item A novel bi-level optimization framework that incorporates LLMs for real-time, language-guided task allocation and planning in adversarial environments.
    \item A hierarchical system design that integrates LLM reasoning with optimizers, supporting constraint enforcement and automatic output validation.
    \item Human interaction for high-level supervision and risk mitigation, validated through comprehensive simulation and hardware experiments.
\end{itemize}

\section{Related Works}

Compared with traditional multi-robot coordination methods, risk-aware and danger-aware strategies have gained significant attention in recent years for their potential to improve resilience and robustness in real-world applications~\cite{majumdar2020should}.
Hazards can cause temporary or permanent damage to robot sensors, disrupt communication links between robots, and degrade the performance of task execution. 
Authors in~\cite{liu2024multi} addressed these dangers in multi-robot target tracking and explored chance-constrained optimization that effectively balances minimizing the risk of failures with maximizing target tracking efficiency.
Resiliency and adaptive strategies with recovery from sensor and communication failures were further extended in~\cite{li2024resilienct}.
To mitigate the influence of sensing and communication attacks, the work in~\cite{abouelyazid2023adversarial} employed neural models for the formation control of the robot team. 
Other methods generated active protections by employing jamming techniques to counter eavesdropping~\cite{10901802}. 
Additionally, active strategies like visual-based anomaly detection~\cite{8963641, sinha2024real} helped ensure safe operations under unforeseen environmental conditions.

However, the environments and adversaries considered in these prior studies are ideal compared to real-world scenarios, which involve challenges such as limited heterogeneous resources, dynamically shifting adversarial zones with changing shapes, and the need for real-time adaptation.
With the increasing complexity of tasks and the hazards of the environments, the problem formulation might not always be feasible, and optimal solutions cannot be obtained in an allowable time with limited computation resources.

Incorporating LLMs into task allocation and motion planning has shown promising results in areas such as reward design~\cite{ma2024eurekahumanlevelrewarddesign}, generating subgoals~\cite{song2023llm, singh2024twostepmultiagenttaskplanning}, and anomaly detection~\cite{sinha2024real}.
While these models can offer valuable strategic input, they are generally ineffective and lack safety guarantees for managing low-level tasks or direct robot actions when used in isolation, as highlighted in previous research~\cite{chen2024solving, li2024challenges, huang2024robotrouting, ji2025genswarmscalablemultirobotcodepolicy}.
Recent work has begun to try to incorporate safety constraints into LLM-based task planning~\cite{10801576, khan2025safetyawaretaskplanning}. 
Furthermore, combining LLMs with conventional optimization techniques can improve both the feasibility and effectiveness of the solutions. 
LLMs can contribute valuable guidance for iterative refinement and problem formulation adjustments~\cite{nie2024importancedirectionalfeedbackllmbased, pmlr-v270-huang25g}, while conventional methods offer stronger guarantees for solution quality and constraint handling.
This integration motivates our work to explore the potential for more robust and reliable performance in multi-agent coordination.

\section{Approach}

\subsection{System Overview}

The proposed framework of multi-robot tracking with hierarchical LLMs \textit{in-the-loop} is shown in Fig. \ref{fig: system}. 
We leverage LLMs in outer and inner feedback loops to iteratively improve the performance for real-time deployment.
The outer loop with task LLM evaluates environment changes and task performance, and ensures that the goal assignments are optimal for the robot team.
The inner loop runs at a high frequency with direct numerical performance feedback to adjust the priority among tracking performance, energy, and safety toward different types of dangers.
With specified variables and parameters, we employ centralized optimizers to generate guaranteed intelligent and safe actions for the robot team.
To address high-level environmental and scenario-specific risks, we incorporate real-time human input, such as feedback on robot performance and environmental hazards, to provide more comprehensive and adaptive responses.
This ensures that the multi-robot team can navigate safely and efficiently in dynamic and potentially hazardous environments.
When the optimized actions are broadcast, the robots execute actions, update their observations, and work together to gather data, detect failures, and provide feedback.

\begin{figure}[!t]
 \vspace{1mm}
    \centering
    \includegraphics[width=1\columnwidth]{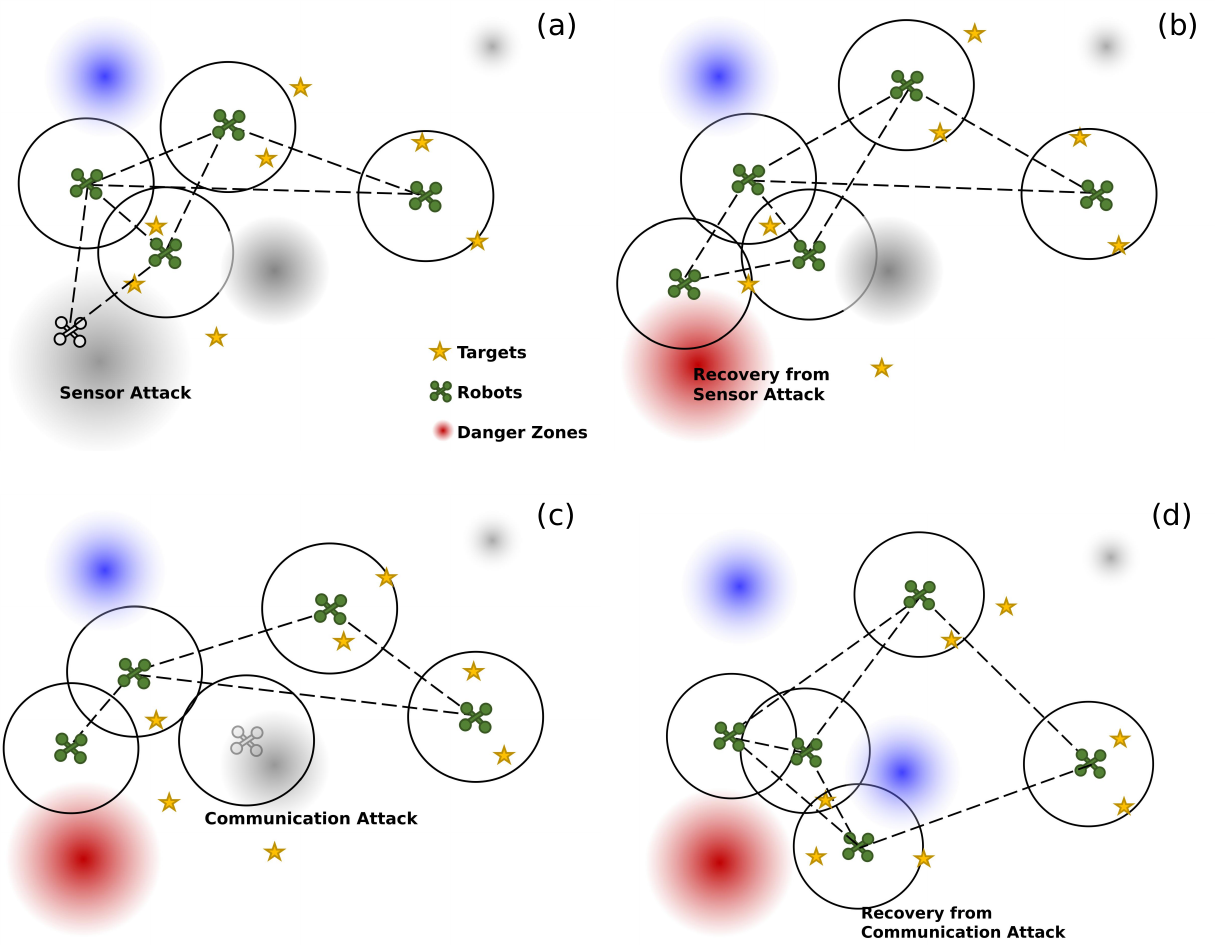}
    \caption{Demonstration of multi-robot target tracking with sensing and communication danger zones. Gray circles indicate undetected zones, red and blue circles are sense and communication zones after detection. The robot team consists of five robots tracking seven targets in an environment with four danger zones. In temporal order, (a–b) show a robot entering a sensing zone, being attacked, and recovering.  (c–d) illustrate a communication failure and subsequent recovery.
    }
    \label{fig: problem}
    \vspace{-5mm}
\end{figure}

\subsection{Task Allocation and Planning as a Bi-level Optimization}

We utilize the multi-robot multi-target tracking problem as an instance of multi-robot coordination with task and motion planning to showcase the capabilities of our framework. 
The unknown hazards in the environment are modeled as danger zones with probabilistic attacks on the robot team.
Following the formulation in~\cite{li2024resilienct}, a team of $M$ sensor-equipped robots is assigned to track $N$ moving targets within environments containing two types of initially unknown hazards: sensing and communication danger zones.
The objective of the multi-target tracking problem is to maximize tracking performance while balancing the deterioration caused by attacks from these danger zones.

This problem requires robots to assign tracking goals toward targets and maximize the tracking performance, which is usually formulated as minimizing the trace of the estimation covariance matrix. 
A demonstration of the tracking process is shown in Fig.~\ref{fig: problem}.
Generally, we formulate this problem of multi-robot coordination in adversarial environments as a constrained multi-objective nonlinear optimization:
\begin{align}
\label{eqn:chance_tracking}
\begin{split}
     \min_{\mathcal{A}, \mathbf{u}} \ \  & \mathcal{J}(\mathcal{A}, \mathbf{u}) = \{ \mathcal{J}_1(\mathcal{A}, \mathbf{u}), \cdots, \mathcal{J}_n(\mathcal{A}, \mathbf{u}) \} ,\\ 
     {s.t.}~ \
    &  \mathcal{H}(\mathcal{A}, \mathbf{u} ) =  \bm{0}, \quad  \mathcal{G}(\mathcal{A},  \mathbf{u} ) \preceq  \bm{0},
\end{split}
\end{align}
where $\mathcal{J}(\cdot)$ represents a set of objectives, such as tracking performance, energy, and safety considerations. 
The term $\mathcal{H}(\cdot)$ denotes the general equality constraint functionals, while $\mathcal{G}(\cdot)$ represents the inequality constraint functionals. 
These constraints encode dynamics-related limitations, feasibility bounds, and other system-specific restrictions.
The variables $\mathcal{A} \in \{ 0, 1\} ^{M \times N}$ represent the task assignment matrix, where $\mathcal{A}_{i,j} = 1$ indicates that robot $i$ is assigned to track target $j$, and $\mathcal{A}_{i,j} = 0$ otherwise. 
The vector $\mathbf{u} = (u_1, \cdots, u_M)     $ denotes the control inputs or general actions for all robots.
The original nonlinear mixed-integer problem is nontrivial and cannot be updated at a higher frequency. 
To improve the efficiency and simplify the reasoning for the LLM, we decouple the problem as a bi-level optimization,  
\begin{subequations}
\begin{align}
    \min_{\mathcal{A}} \quad & \mathcal{J}(\mathcal{A}, \mathbf{u}^*) ,\\
    \text{s.t.} \quad & \mathbf{u}^*(\mathcal{A}) \in \arg\min_{\mathbf{u}} \left\{\mathcal{J}(\mathcal{A}, \mathbf{u}) \mid \right. \\
    & \qquad \qquad \mathcal{H}(\mathcal{A} , \mathbf{u}) = \bm{0}, \ \mathcal{G}(\mathcal{A}, \mathbf{u}) \preceq \bm{0} \}, \\
    & \mathcal{H}(\mathcal{A}, \mathbf{u}^*) = \bm{0}, \\
    & \mathcal{G}(\mathcal{A}, \mathbf{u}^*) \preceq \bm{0}.
\end{align}
\end{subequations}
The high-level problem usually evaluates the task performance based on the cost of deployments towards robots, as discussed in~\cite{9349130}.
The subproblem focuses on optimizing the robots' control commands, given the assignment matrix generated by the high-level problem.
Hence, the low-level  optimization can be formulated as
\begin{subequations} 
    \label{eq:optimization} 
    \begin{align} 
        \min_{\mathbf{u}, \bm{\nu}, \bm{\xi} }& \ w_1 f(\mathbf{x}_{t+1}) + w_2 \lVert \mathbf{u}\rVert + w_3 \lVert \bm{\nu} \rVert + w_4 \lVert \bm{\xi} \rVert,  \\
        \textrm{s.t.} & \quad \mathbf{x}_{t+1} = g(\mathbf{x}_{t}, \mathbf{u}) , \quad \label{eq: dynamics} \\
        & \quad \mathcal{G}(\mathbf{x}_{t+1}, \mathcal{S}) \preceq \bm{\nu} ,\label{eq: sensing} \\
        & \quad \mathcal{G}(\mathbf{x}_{t+1}, \mathcal{C}) \preceq \bm{\xi} ,\label{eq: communication} \\
         & \quad \mathbf{u} \in \mathcal{U}, \quad \mathbf{x}_{t+1} \in \mathcal{X}_{t+1}^{\text{free}}, \quad  \bm{\nu}, \bm{\xi} \succeq  \mathbf{0},
    \end{align} 
\end{subequations}
where the objective function is composed of four components weighted by $ w_1, w_2, w_3, w_4 \in \mathbb{R}_+$, $f$ represents tracking performance.
The equation (\ref{eq: dynamics}) represents the dynamic model of robots, and $\mathbf{x}_{t}$ is the robot state at time $t$.
The constraint (\ref{eq: sensing}) denotes the probabilistic safe constraints toward known sensing danger zones $\mathcal{S} = (\mathcal{S}_1, \cdots, \mathcal{S}_p)$, and (\ref{eq: communication}) represents the safety conditions for known communication danger zones $\mathcal{C} = (\mathcal{C}_1, \cdots, \mathcal{C}_q)$ with slack variables $\bm{\nu}, \bm{\xi} $ for relaxation.

Instead of deploying an LLM for each individual agent, we adopt a partially centralized optimization approach, solved with solvers such as FORCES Pro~\cite{FORCESPro}. 
In this framework, a task LLM guides task assignment, while an action LLM assists in tracking processes. 
The prompt designs and verification methods will be discussed in the following sections.

\subsection{Task LLM for Reconfiguration}

We query the Task LLM to directly generate the assignment matrix, avoiding explicit optimization with matrix-based hard constraints, which is not straightforward to reason.
Task assignment in a multi-robot tracking problem requires comprehensive information about robot capabilities, available resources, and the current status of both robots and dynamic tasks (targets). 
Additionally, historical data is important to evaluate whether a robot's assignment should be updated.
The following information is provided to the task LLM in a descriptive language format: (1) \textbf{robot and target status}: This includes the dynamic states of robots (e.g., position, velocity), target states, and the current status of robots failures by attacks (e.g., whether they are affected by sensing or communication danger zones). (2) \textbf{environmental status}: Known locations of sensing and communication danger zones, along with their impact on the robots. (3) \textbf{robot capacities}: The number of targets a robot can track at once, which may also be dynamically adjusted based on current conditions.  (4) \textbf{historical data}: The most recent $k$ results about states of robots and targets, information on previous danger zone attacks, and prior task assignments.
This historical context helps the LLM make informed decisions about potential reassignment needs.

\subsubsection{Prompt Design}
The system prompt is designed to feed into LLM by clearly defining its role as an optimization engine tasked with assigning tracking targets to robots. 
\llmbox{
You are an optimizer with the goal of assigning robots to track targets. 
}
The user input provides essential real-time and historical information about robots, targets, and the environment. An example is provided:

\llmbox{The recent k results of status and observations are ... The \textcolor{blue}{[]}th information is as follows.
Drones are currently at the following positions: Drone \textcolor{blue}{[ID]} is at \textcolor{blue}{[]} .... Targets are currently at the following positions: Target \textcolor{blue}{[ID]} is at \textcolor{blue}{[]} ... The known sensing zones are: Drone \textcolor{blue}{[ID]} knows the sensor zone \textcolor{blue}{[ID]} located at ... No attack has been detected. The last trace of the tracking estimation covariance matrix is \textcolor{blue}{[]}. }
The optimization constraints and request for LLM outputs are described in the form of 

\llmbox{
Each drone has the ability to track at most \textcolor{blue}{[]} targets, and each target should be tracked by at least one drone as possible. Please provide the new tracking assignment for each drone in each line with ''. 
}
This structured approach ensures that the LLM has all the necessary data, constraints, and output expectations.

\subsubsection{Output Verification}

We employ two types of checkers: the first ensures that the LLM output adheres to the required format with numerical data, while the second verifies whether the output meets the specified hard constraints.
The verification function is designed to ensure that the output from the LLM can be used effectively for low-level optimization.
It helps to determine whether the current output should be used or skipped, adjusting the prompt for the next query if needed.
The assignment matrix is evaluated using
\begin{align}
\beta(\mathcal{A}) = \mathbf{1}^T \max( \mathbf{0}, \mathcal{A} \mathbf{1} - \mathbf{c}) + \mathbf{1}^T \max( \mathbf{0}, \mathbf{1} - \mathcal{A}^T \mathbf{1} ),
\end{align}
where $\mathbf{c} = (c_1, \cdots, c_M) $ is the vector of robot capacities, \(\mathbf{1}\) represent a vector of ones.
The first term evaluates the robot capacity constraint as satisfying, where $c_i$ is the maximum number of targets the robot $i$ can track. 
The second term qualifies as each target being tracked by at least one robot.
If $\beta(\mathcal{A}) > 0 $, then the constraints are not met, and we will keep the previous assignment and discard the current one.

\subsection{Action LLM for Performance Adaptation}

\subsubsection{Prompt Design}

We specify the reasoning meaning of different objectives to the action LLM to evaluate their importance and adjust weights $(w_1, w_2, w_3, w_4)$ accordingly. 
The description for each objective is:
\llmbox{
\begin{itemize}
    \item[1.] tracking error computed by the trace of the estimation covariance matrix of the targets
    \item[2.] control cost computed by the norm of control input
    \item[3.] slack variables of safety constraints to avoid sensing danger zones
    \item[4.] slack variables of safety constraints to avoid communication danger zones
\end{itemize}
}
We use action LLM to evaluate the relative importance of each objective and adjust the parameters $w_1, w_2, w_3, w_4$ accordingly.
The weight adjustment is a short-horizon decision. 
Therefore, we provide the LLM with the current status of robots, targets, and environments, with the last weight values. 
Similarly, the system prompt is designed as:

\llmbox{You are a multiple objective optimizer with the goal of specifying the weights of each objective function.}
In addition to status, the weight is provided with the specified output format.

\llmbox{The current weights for objective functions are: ... You should give a new weight as a list with a length of 4.}

\subsubsection{Output Verification}

The weights in our optimization framework typically indirectly influence the results. 
To determine whether the output generated by the LLM is appropriate for optimization, we add empirical bounds for each weight. 
We use such bounds to filter the LLM's results, ensuring that only outputs meeting predefined criteria are passed on for further optimization.

\vspace{-0.1cm}
\subsection{Human Input}

We consider human reactions to supervise the system to inspect abnormal behaviors, risks, and uncertainty.
Though humans are generally not able to directly manage massive numerical data in real-time, like the exact positions of all robots, targets, and danger zones, but can have a sense of the actions of robots given some visualizations of data and real-world unmodeled conditions.
Human input is optional and added to the user prompt if the human has comments about the current performance.

We list the following types of input 
(1) \textbf{performance qualifications}: Humans can provide questions or concerns about the general performance. For example, humans can specify one robot to conduct subtraction, like focusing on the track part of the targets. 
Humans can also raise questions about the performance, reporting the missing tracking of some targets, zigzag actions of robots, etc. 
(2) \textbf{risk assessment}: For environmental hazards, humans can specify some high-level evaluation, such as pushing the robot team to be more conservative, or detailed information, such as avoiding a particular danger region if the target is very close to this region. 
(3) \textbf{abnormal specifies}: When deploying the framework to hardware platforms, sometimes there are extra factors that might affect the performance. For example, a robot might get trapped in a local minimum and not move.
Then, a human can specify that the robot doesn't move for a time and what's the reason for that. 
In addition, the danger zone we assume might be more severe than its model and thus needs priority adjustment for safety.

\section{Results}

This section presents a series of simulations and real-world experiments designed to evaluate the following hypotheses:
\begin{itemize}
    \item[\textbf{H1.}]  LLMs are able to reconfigure the task allocation of robots based on the structured language feedback of robots and environment information. 
    \item[\textbf{H2.}]  LLMs are able to adjust the objective weights based on the text feedback.
    \item[\textbf{H3.}]  Integrating LLMs into the system can boost tracking task performance and reduce risks.
    \item[\textbf{H4.}]  The proposed LLM-in-the-loop framework is deployable in real-time in robotic hardware.
\end{itemize}

\subsection{Simulation Evaluation}

\subsubsection{Implementation Details}

We evaluated the capabilities of our framework through numerical simulations in different scenarios. 
For our experiments, we utilized GPT-4o~\cite{achiam2023gpt}, GPT-4.1-mini, LLaMA-3-70B~\cite{grattafiori2024llama3herdmodels}, and DeepSeek-V3~\cite{deepseekai2025deepseekv3technicalreport}, setting the temperature to 0 to ensure deterministic outputs.
The token limit is $50 \times (\bar{c} + 2)$, where $\bar{c}$ is the average robot tracking capacity. 
GPT-4o gets extra tokens for its detailed outputs.
During the simulations, the inner LLM was updated every 2-3 steps to manage tuned performance, while the outer LLM was refreshed every 8-10 steps to handle higher-level task assignments. 
The simulation runs for 300 steps.

\begin{figure}[!t]
    \centering
    \includegraphics[width=1\columnwidth]{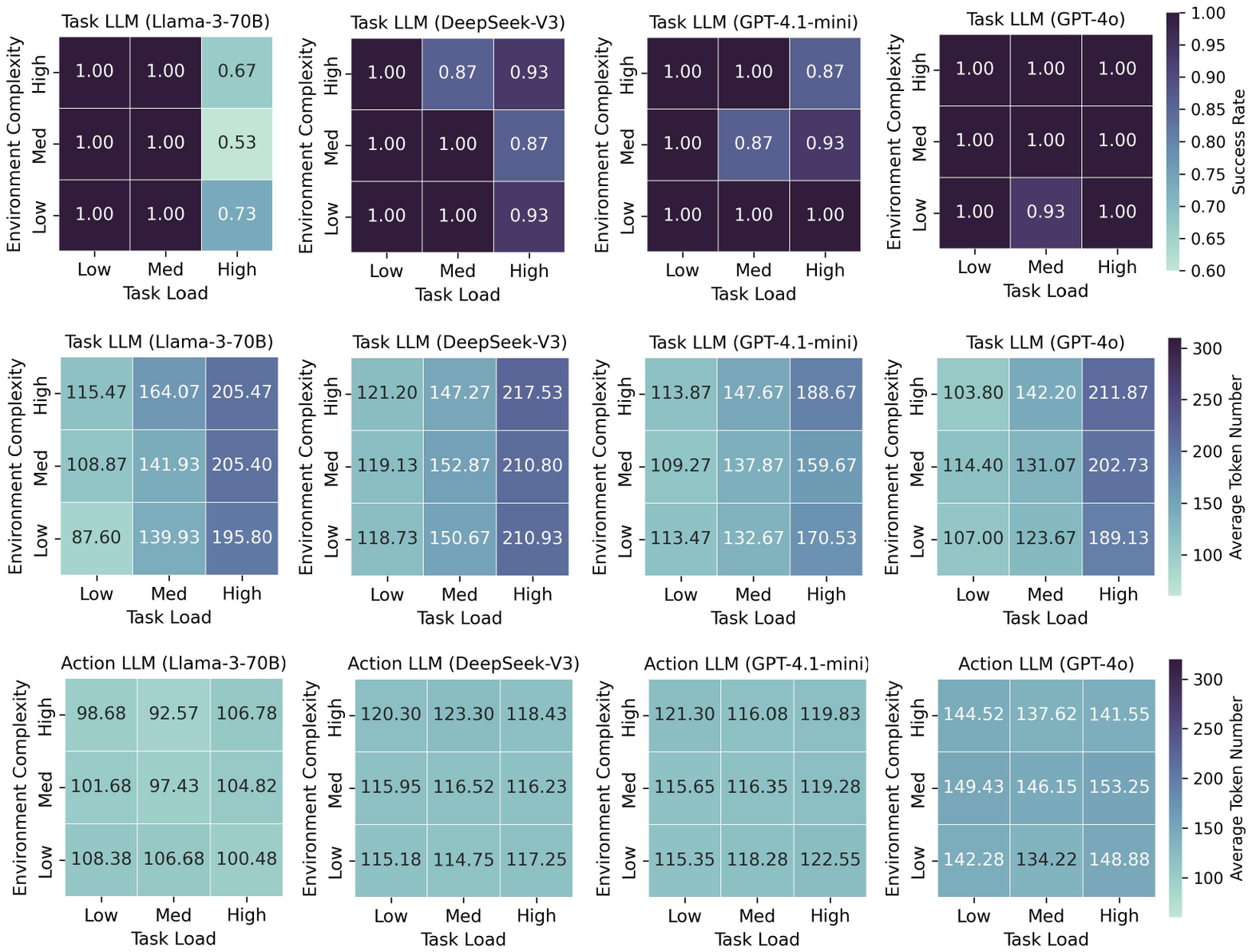}
     \vspace{-4mm}
     \caption{Success rate and token count generated by different models for the Task LLM (a) and Action LLM (b), evaluated under varying task loads and environmental complexities.}
    \vspace{-5mm}
    \label{fig: token}
\end{figure}

\subsubsection{Reasoning Ability}

We design scenarios with varying levels of complexity to evaluate the performance of our framework, focusing on two key dimensions: task load, quantified by the robot-to-target ratio, and environmental complexity, measured by the number and distribution of danger zones.
The task load is categorized into three levels as underloaded (0.5), balanced (1.0), and overloaded (2.0) based on the ratio of targets to robots. 
For example, with four robots, these levels correspond to tracking 2, 4, and 8 targets, respectively, assuming each robot can track up to 3 targets.
Environmental complexity is similarly categorized into three levels, with 2, 4, and 8 danger zones representing low, medium, and high complexity, respectively. 
This design allows us to isolate and analyze the impact of internal resource limits and external environmental stressors on tracking performance, robustness, and real-time feasibility.

We evaluated four LLMs across the different scenarios, with Fig.~\ref{fig: token} summarizing success rates and token usage for both the Task and Action LLMs.
The success rate is evaluated in the output verification module, and token usage reflects the system’s real-time ability and scalability.
All models achieved a 100\% success rate for the Action LLM,  
For the Task LLM, success rates are shown in the first row of the figure.
The token count for the Task LLM generally increases with the number of robots and targets, reflecting the growing complexity of task assignments. 
In contrast, the Action LLM’s token usage remains relatively stable, as the output complexity does not scale significantly with the prompt size.
We also tested the Task LLM with and without prior examples of task assignments. 
Even without such references, the models consistently generated feasible outputs that satisfied all constraints, demonstrating strong generalization and adaptability under increasing task complexity.
These results demonstrate that LLMs can effectively generate task assignments (\textbf{H1}) and dynamically adjust objective weights (\textbf{H2}) based on structured language inputs.

\subsubsection{Ablation Study}

We compare three approaches: a baseline system using expert-defined parameters and fixed task allocation without any LLM (w/o LLMs), one with LLMs in the loop (w/ LLMs), and one with LLMs and human input into the system (w/ LLMs + Human).
We consider a scenario with two drones tracking four targets in the same direction. 
We specify targets to have different velocities to increase the complexity of the conditions. 
The environment includes two sensing danger zones and one communication danger zone.
Human input is provided at a lower frequency to encourage the LLMs to independently improve their performance through adaptation. An example prompt is:

\llmbox{Focus more on tracking targets; The trace is not good.}

To illustrate, we use GPT-4o with the results presented in Tab. \ref{tab: ablation}.
By adding the LLMs in the loop to adjust the parameters, the system can achieve better tracking performance and get fewer attacks from sensing and communication danger zones (\textbf{H3}).
With human input to express concern about the performance, LLMs can dramatically reduce tracking errors computed by the trace of the estimation covariance matrix. 
\begin{table}[!th]
    \renewcommand\arraystretch{1.1}
    \setlength{\tabcolsep}{5pt}
     \vspace{0.2cm}
    \centering
    \caption{Comparison of Overall Performance \label{tab: ablation}}
     \vspace{-0.1cm}
    \begin{tabular}{ccccc}
    \hline 
     & \tabincell{c}{\textbf{Accumulated} \\ \textbf{Trace} }  &  
   \tabincell{c}{\textbf{Sensing} \\ \textbf{Attacks} } &
   \tabincell{c}{\textbf{Comm.} \\ \textbf{Attacks} } &
   \tabincell{c}{\textbf{Trajectory } \\ \textbf{Length (m) }} \\
   \hline 
   w/o LLMs   & 120.08 & 31 & 11 & 17.53   \\
    w/ LLMs   & 116.86 & 23 & 10 & 17.69 \\
    w/ LLMs + Human  & 100.98 & 19 & 7  & 18.55 \\
    \hline 
    \end{tabular}
    \vspace{-0.3cm}
\end{table}

\subsection{Hardware Experiments}

\begin{figure}[!t]
\vspace{0.1cm}
    \textbf{}
    \centering
    \subfigure[Outdoor]{
\includegraphics[height = 2.9cm]{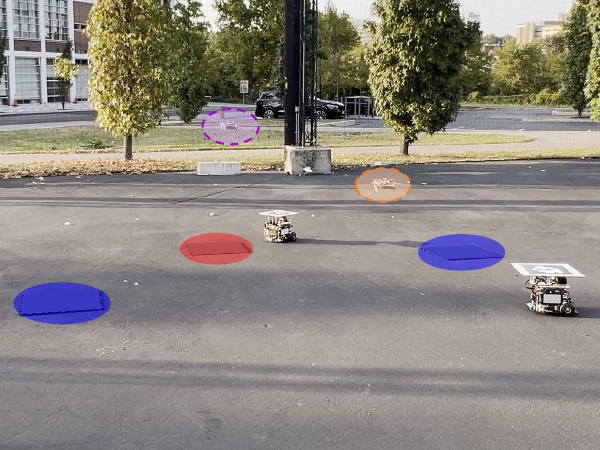}
    }
    \subfigure[Indoor]{
\includegraphics[height = 2.9cm]{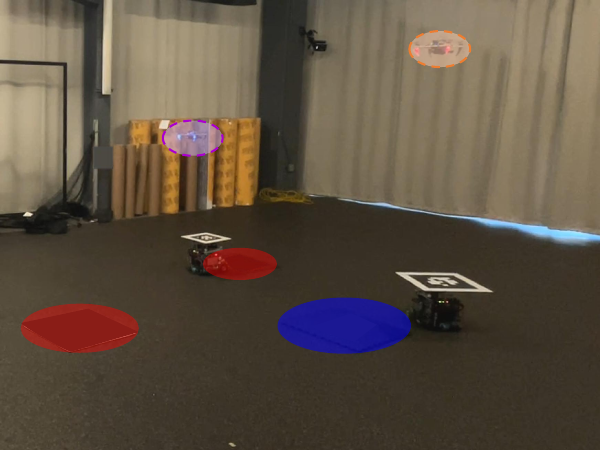}
    }
     \vspace{-0.2cm}
    \caption{Hardware experiment of multi-drone target tracking with various danger zones (red disks represent sensing danger zones, and blue disks represent communication danger zones). }
    \label{fig: exp}
    \vspace{-0.5cm}
\end{figure}

We validated the ability of the proposed framework through experiments with a team of four robots in both indoor and outdoor scenarios (\textbf{H4}), as shown in Fig.~\ref{fig: fig1} and~\ref{fig: exp}.
The outdoor experiments include more challenges like noisy detection, estimation, and control under disturbance.
We set up two ground robots to serve as moving targets, with each carrying an AprilTag~\cite{5979561} on top. 
In hardware experiments, the update frequency of each module was adjusted based on the specific scenario and runtime constraints. For example, in a real-world deployment involving two aerial drones tracking two ground robots, the computation time and update frequency of each major module are summarized in Tab.~\ref{tab:runtime}.  
The action LLM runs 4–5 times more frequently than the task LLM, enabling more responsive low-level adaptations.
To support long-horizon reasoning, the task LLM received a history of robot and environment states from the past 5 steps. 
If the LLM output is invalid or violates constraints, the system retries with adjusted prompts until a valid response is obtained, ensuring robustness to occasional failures.

To effectively detect and track the moving targets, two drones were employed. 
Each drone is equipped with a downward-facing VGA camera for state estimation and target detection. 
An Unscented Kalman Filter is running on board the drone to fuse the 30Hz VIO odometry with the IMU to get 150Hz pose information. 
We used an additional AprilTag to create a common reference frame for all drones. 
Once an aerial agent observes the tag, the agent estimates the relative transformation between the common reference frame and its odometry frame and registers the detected target in the common reference frame. 
The robot receives global waypoints from the ground station and transforms them into its odometry frame for trajectory generation and control. 
The state estimation, tag detection, trajectory planning, and controller run onboard the drones. 
\begin{table}[!t]
\vspace{0.1cm}
    \renewcommand\arraystretch{1.1}
    \setlength{\tabcolsep}{5pt}
    \centering
    \caption{Module Execution During a Single Experiment Trial}
    \begin{tabular}{lccc}
        \hline
        \textbf{Module} & \textbf{Frequency (Hz)} & \textbf{Avg. Runtime (s)} & \textbf{Success Rate} \\
        \hline
        Task LLM     & 0.05   & 1.3639   & 100\% \\
        Action LLM   & 0.2    & 0.8911   & 100\% \\
        Optimizer    & 10     & 0.0034   & 100\% \\
        \hline
    \end{tabular}
    \label{tab:runtime}
    \vspace{-0.5cm}
\end{table}
We created a series of human inputs during the experiments to see how they would impact the systems and help the LLMs better understand real-world conditions. 
These inputs focus on both performance and environmental concerns, allowing us to test the system's responsiveness and adaptability.
For performance input, we provide some examples of human expression about tracking performance, overall strategy, and individual robot behavior.

\llmbox{
    \begin{itemize}
        \item The robots should be more aggressive. 
        \item Why does the Drone 21 get stuck? % why does the Drone 21 look like it's stuck? or Why does the Drone 21 get stuck?
        \item The Drone 21 should focus on tracking target 122.
    \end{itemize}
}

To address unexpected changes regarding environmental conditions or hazards, the following examples are provided:

\llmbox{
    \begin{itemize}
        \item The sensing danger zone causes more severe damage than we assume.
        \item The target 122 is in the danger zone.
    \end{itemize}
}

In addition, a case study is shown below.
Initially, the output specified tracking both targets, as follows:

\llmoutput{
\begin{itemize}
 \item[1.] Drone 21 should track Target 122 and Target 124.
\item[2.] Drone 26 should track Target 122 and Target 124.
\end{itemize}
}

However, tracking both targets simultaneously can lead to ambiguous behavior, making it unclear whether the drones are effectively monitoring the targets. 
Consequently, the human supervisor raised a concern,

\llmbox{
The human supervisor has inputs: The robots do not seem to track the targets.
}

In response, the Task LLM reassigned and refocused each drone to a single target. 
The updated plan was:

\llmoutput{
Based on the information provided, here is the new tracking assignment for each drone:\
  - Drone 21 will track Target 122\\
  - Drone 26 will track Target 124\\
  Feedback to the human supervisor: The new tracking assignment has been optimized based on the capabilities and constraints of the drones, ensuring that each target is being tracked by at least one drone. The assignments have been made considering the capabilities and sensor failures of the drones.
}

This interaction process demonstrates how the system uses human input to reconfigure task allocations, improving real-time decision-making and operational efficiency.

\section{Conclusion} 
\label{sec:conclusion}

In this work, we propose a hierarchical framework that combines the reasoning capabilities of LLMs with optimization techniques for multi-robot target tracking. 
Our results demonstrate the capacity of this framework to handle unexpected conditions, showing improved performance in resilience, adaptability, and tracking accuracy, particularly in dynamic and hazardous environments.
This highlights the potential of LLM-integrated robotics systems in a wide range of real-world applications.

This work explores a promising direction for optimization-informed LLMs, but also presents limitations. 
LLM inference introduces latency and relies on carefully crafted prompts, structured optimization formulations, and constrained outputs, limiting scalability to large-scale reactive multi-robot systems.
Future work will focus on improving scalability and robustness while maintaining lightweight onboard deployment and incorporating fine-tuning with human feedback.

\vspace{-0.2cm}

\bibliography{references}
\end{document}